\pdfoutput=1

\documentclass[11pt]{article}

\usepackage{ACL2023}

\usepackage{times}
\usepackage{latexsym}

\usepackage[T1]{fontenc}

\usepackage[utf8]{inputenc}

\usepackage{microtype}

\usepackage{inconsolata}
\usepackage{booktabs}
\usepackage{graphicx}
\usepackage{amsmath}

\newcommand{\ourM}{MolXPT}

\newcommand{\som}{$\langle$som$\rangle$}
\newcommand{\eom}{$\langle$eom$\rangle$}

\title{\ourM{}: Wrapping Molecules with Text for Generative Pre-training}

\author{Zequn Liu$^1$\thanks{$^*$Equal contribution. This work was done when Z. Liu and W. Zhang were interns at Microsoft Research AI4Science.}, Wei Zhang$^{2\,*}$, Yingce Xia$^3$\thanks{$^\dag$Corresponding authors.}\;, Lijun Wu$^3$, Shufang Xie$^4$, \\\textbf{Tao Qin$^3$, Ming Zhang$^{1\,\dag}$ and Tie-Yan Liu$^3$}\\
$^1$ Peking University; $^2$ University of Science and Technology of China \\
$^3$ Microsoft Research AI4Science; $^4$ Renmin University of China\\
   \texttt{\{zequnliu,mzhang\_cs\}@pku.edu.cn}; 
  \texttt{weizhang\_cs@mail.ustc.edu.cn} \\
  \texttt{\{yingce.xia, lijunwu, taoqin, tyliu\}@microsoft.com}\\
  \texttt{shufangxie@ruc.edu.cn}}

\begin{document}
\maketitle
\begin{abstract}
Generative pre-trained Transformer (GPT) has demonstrates its great success in natural language processing and related techniques have been adapted into molecular modeling. Considering that text is the most important record for scientific discovery, in this paper, we propose \ourM{}, a unified language model of text and molecules pre-trained on SMILES (a sequence representation of molecules) wrapped by text. Briefly, we detect the molecule names in each sequence and replace them to the corresponding SMILES. In this way, the SMILES could leverage the information from surrounding text, and vice versa. The above wrapped sequences,  text sequences from PubMed and  SMILES sequences from PubChem are all fed into a language model for pre-training. 
Experimental results demonstrate that \ourM{} outperforms strong baselines of molecular property prediction  on MoleculeNet, performs comparably to the best model in text-molecule translation while using less than half of its parameters, and enables zero-shot molecular generation without finetuning. 
\end{abstract}

\section{Introduction}
Generative pre-trained Transformer (GPT), like GPT-3 \citep{GPT3} and ChatGPT \citep{chatGPT}, have obtained great success in natural language processing. 
They usually have billions of parameters and are trained on large corpus \cite{taylor2022galactica,singhal2022large}. By witnessing their great power, people start transferring language models to chemical \citep{MolGPT} and biological domains \citep{protGPT2}. For example, a small molecule (e.g., an oral drug) can be represented using simplified molecular-input line-entry system (SMILES) \cite{smiles}, which is a sequence obtained by traversing the molecular graph using depth-first-search and several rules for branching, aromaticity, etc. After serializing molecules, people pre-train language models on SMILES \citep{MolGPT,tong2021generative,neuralScale} and obtain promising results for molecular generation.

Text is the most important record for molecular science and more generally, scientific discovery \cite{beltagy-etal-2019-scibert}. It describes detailed properties of molecules, like how to synthesize the molecule \cite{feng2016sulfur},  whether the molecule is toxic \cite{juurlink2003drug}, etc. BioGPT \citep{bioGPT} and PubMedGPT \citep{pubmedGPT} are two language  models trained on biomedical literature. Recently, a new trend is to jointly model SMILES and scientific text so as to obtain  shared representations across the two modalities. MolT5 is a T5-like \citep{2020t5} model, where several spans of the text/SMILES are masked in the encoder and they should be reconstructed in the decoder. Galactica \citep{taylor2022galactica} is a GPT-like \cite{GPT3} model  pre-trained on various types of inputs, like text, SMILES, protein sequences, etc. Although those models demonstrate progress in prediction and generation tasks, they do not explicitly leverage the relation between molecules and text. An intuition is that, in scientific literature, when a molecule name appears in a sentence, the surrounding context could be a description of the molecule. This should be useful information for joint training but is ignored in those models. 

\begin{figure*}[!t]
\centering
\includegraphics[width=\textwidth]{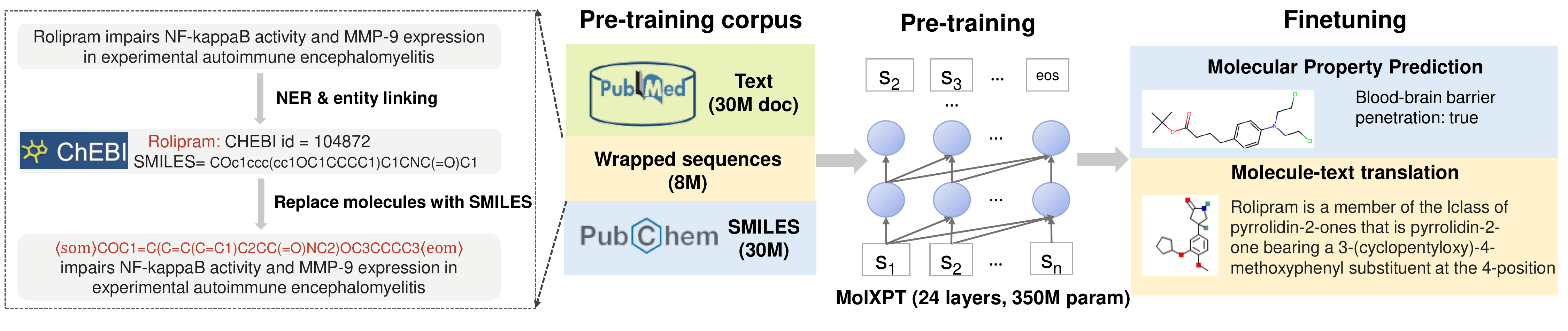}
\caption{Framework of \ourM{}. \ourM{} is pretrained on text from PubMed, SMILES from PubChem and wrapped sequences between SMILES and text. The wrapped sequences are obtained by applying NER and entity linking to text and then replacing matched molecular mentions with SMILES. \ourM{} can be finetuned for various text and molecular downstream tasks, like molecular property prediction and molecule-text translation.}
\label{flowchart}
\end{figure*}

To leverage such relations, in this work, we propose a novel molecule-text language model (\ourM{}), which is trained on ``wrapped'' sequences: Given a sentence, we detect the molecular names with named entity recognition tools, and if any, replace them to the corresponding SMILES and obtain the ``wrapped'' sequence between SMILES and text. We pre-train a 24-layer \ourM{} (with $350M$ parameters) on 8M wrapped sequences, as well as 30M SMILES from PubChem \citep{pubchem} and 30M titles and abstracts from PubMed (a popular biomedical literature search engine).

After pre-training, we finetune \ourM{} on MoleculeNet (a benchmark about molecular property prediction) \citep{moleculenet} and molecule-text translation \citep{molt5} using prompt-based finetuning. On MoleculeNet, \ourM{}  outperforms strong baselines with sophisticated design like GEM \citep{fang2022geometry}. On text-molecule translation, \ourM{} performs comparably with the state-of-the-art model, MolT5-large \cite{molt5}. MolT5-large has $800M$ parameters while \ourM{} only uses 44\% of its parameters.  We also verify that \ourM{} has the zero-shot ability on text-to-molecule generation. 

\section{Our Method}

\ourM{} is a language model  pre-trained on heterogeneous data including scientific text, SMILES sequences, and ``wrapped'' sequences between SMILES and text. Due to the flexible input, we can finetune it for various text and molecular tasks. The framework of \ourM{} is in Figure \ref{flowchart}.
\subsection{Pre-training corpus}
For scientific text, we use the titles and abstracts of 30M papers from PubMed\footnote{\url{https://ftp.ncbi.nlm.nih.gov/pubmed/}}. For molecular SMILES, we randomly choose 30M molecules from PubChem\footnote{\url{https://pubchem.ncbi.nlm.nih.gov/}} \citep{pubchem}.

The wrapped sequences are constructed via a ``detect and replace'' pipeline. We first use BERN2 \cite{bern2}, a widely used named entity recognition (NER) tool for biomedical purpose, to detect all mentions of molecules and link them to the entities in public knowledge bases like ChEBI \cite{chebi}. After that, we can retrieve the molecular SMILES of the matched entities. Finally, we replace the molecular mentions to their corresponding SMILES. An example is shown in the left panel of Figure \ref{flowchart}.
The wrapped sequences must contain at least one molecular SMILES. We eventually obtain 8M wrapped sequences in total.

Text and SMILES  are tokenized separately. For  text, we use byte-pair encoding (BPE) \citep{sennrich2016BPE} to split the words into subwords. The number of BPE merge operation is $40$k.  For SMILES sequences (including those in wrapped sequences), we tokenize them  with the regular expression from \citep{schwaller2018found}. For each SMILES sequence $S$, we add a start-of-molecule token \som{} at the beginning of $S$ and append an end-of-molecule token \eom{} at the end of $S$.


\subsection{Model and training}
\noindent{\em Model architecture}: \ourM{} has the same architecture as the GPT models \citep{gpt2}. Due to computational resource limitation, in this paper, we follow the GPT-2$_{\rm medium}$ configuration  with 24 layers, 1024 hidden size and 16 attention heads. The maximum length of input we can process is 2048 and the vocabulary size is 44536. In total, our model has 350M parameters.

\begin{table*}[!htbp]
\centering
\resizebox{0.9\textwidth}{!}{
\begin{tabular}{lcccccccccccccc}
\toprule
\textbf{Dataset} & BBBP & Tox21 & ClinTox & HIV & BACE & SIDER & Avg \\
\textbf{\#Molecules} & 2039 & 7831 & 1478 & 41127 & 1513 & 1478 & \\
\midrule
G-Contextual& $70.3\pm1.6$ & $75.2\pm0.3$ & $59.9\pm8.2$ & $75.9\pm0.9$ & $79.2\pm0.3$ & $58.4\pm 0.6$ & $69.8$ \\
G-Motif &$66.4\pm3.4$&$73.2\pm0.8$&$77.8\pm2.0$&$73.8\pm1.4$&$73.4\pm4.0$&$60.6\pm1.1$ & $70.9$\\
GROVER$_{\rm base}$&$70.0\pm0.1$&$74.3\pm0.1$&$81.2\pm3.0$&$62.5\pm0.9$&$82.6\pm0.7$&$64.8\pm0.6$&$72.6$ \\
GROVER$_{\rm large}$&$69.5\pm0.1$&$73.5\pm0.1$&$76.2\pm3.7$&$68.2\pm1.1$&$81.0\pm1.4$&$65.4\pm0.1$&$72.3$\\
GraphMVP &$72.4\pm1.6$&$75.9\pm0.5$&$79.1\pm2.8$&$77.0\pm1.2$&$81.2\pm0.9$&$63.9\pm1.2$& $74.9$\\
MGSSL& $70.5\pm1.1$ & $76.5\pm0.3$&$80.7\pm2.1$&$79.5\pm1.1$& $79.7\pm0.8$&$61.8\pm0.8$ & $74.8$\\
GEM & $72.4\pm0.4$ & \textbf{78.1 $\pm$ 0.1} & $90.1\pm1.3$ & \textbf{80.6 $\pm$ 0.9} & $85.6\pm1.1$ & $67.2\pm0.4$ & $79.0$ \\
\hline
KV-PLM &  $74.6\pm0.9$ & $72.7\pm0.6$ & -- & $74.0\pm1.2$ & -- & $61.5\pm1.5$ & --\\
Galactica& $66.1$ & $68.9$ & $82.6$ & $74.5$ & $61.7$ & $63.2$ & $69.5$ \\
MoMu & $70.5\pm2.0$ & $75.6\pm0.3$ & $79.9\pm4.1$ & $76.2\pm0.9$  & $77.1\pm1.4$ & $60.5\pm0.9$ & $73.3$ \\
\midrule
\ourM{} & \textbf{80.0 $\pm$ 0.5} &  $77.1\pm0.2$  & \textbf{95.3 $\pm$ 0.2} & $78.1\pm0.4$ & \textbf{88.4 $\pm$ 1.0} & \textbf{71.7 $\pm$ 0.2} & \textbf{81.9} \\
\bottomrule
\end{tabular}}
\caption{Results on MoleculeNet. The evaluation metric is ROC-AUC. Bold fonts indicate the best results.}
\label{tab:moleculenet}
\end{table*}

\noindent{\em Pre-training}: 
The pre-training objective function of \ourM{} is the  negative log-likelihood. Mathematically, let $\mathcal{D}=\{x_i\}_i$ denote the collection of sequences of the three types of the data, and $x_i=(s_{i,1},s_{i,2},\cdots,s_{i,n_i})$ is the $i$-th sequence with $n_i$ tokens. The training objective function is:
\begin{equation*}
\min -\frac{1}{|\mathcal{D}|}\sum_{i=1}^{\vert\mathcal{D}\vert}\sum_{j=1}^{n_i}\log P(s_{i,j}|s_{i,j-1},s_{i,j-2},\cdots,s_1). 
\end{equation*}
The pre-training details are  left in Appendix \ref{app:pretrain_details}.

\noindent{\em Prompt-based finetuning}: \ourM{} can be finetuned for downstream tasks about molecules and text. Adding classification or regression heads to pre-trained backbone models introduces the gap between pre-training and finetuning \cite{GPT3,chen2022adaprompt,gu2022ppt}. Therefore, we adopt prompt-based finetuning \cite{lmbfs} to unify different tasks into a sequence generation task, which is consistent with the pre-training objective. Briefly, given a task, we convert the input and output into text and/or SMILES sequences, equip the sequences with task-specific prompts and finetune using language modeling loss. Prompts for MoleculeNet and text-molecule translation are introduced in the Section \ref{sec:pubmedqa_exp} and \ref{sec:textmol_trans} respectively.


\noindent{\em Discussion}: Some  works also try to jointly model text and molecules. \citet{Zeng2022KV-PLM} propose KV-PLM, where SMILES sequences are appended after molecule names for pre-training. \citet{su2022molecular} use contrastive learning between  text and molecular graphs. Our \ourM{} is a generative model while the above two models are not. Both of them are built upon SciBERT \citep{beltagy-etal-2019-scibert}, a BERT model \cite{devlin-etal-2019-bert} for scientific literature. \ourM{} is complementary to them.

\section{Experiments}
We evaluated \ourM{} on two downstream tasks: (1) molecular property prediction on MoleculeNet \cite{moleculenet}, which is to predict whether the given molecule has specific properties;
(2) the generation between text descriptions and molecules \cite{molt5}, where both molecules and text should be considered. In this section, we focus on introducing task definition, prompt design and results while leaving the detailed finetuning hyper-parameters in Appendix \ref{sec:implementation}.

\subsection{Results on MoleculeNet}
\label{sec:pubmedqa_exp}

MoleculeNet \cite{moleculenet} is a widely-used benchmark for molecular modeling, which has more than 700$k$ compounds for various different properties. We choose six molecular classification tasks for evaluation, which are BBBP, Tox21, ClinTox, HIV, BACE and SIDER. Details are left in Appendix \ref{sec:moleculenet_baseline}.
We follow GEM \citep{fang2022geometry} to split the data into training/validation/test sets based on the scaffold. For these tasks, the input is a SMILES and the output is a binary label. 

\noindent{\em Finetuning strategy}: Previous molecular property prediction models mainly use SMILES sequences or molecular graphs as input, while we can use the ``wrapped'' sequences. For example, one task is to predict the blood-brain barrier penetration (BBBP) of a molecule. Therefore, the prompt is ``{\em We can conclude that the BBB penetration of }\som{} $\langle$\texttt{SMILES}$\rangle$ \eom{} is $\langle$tag$\rangle$'', where $\langle$\texttt{SMILES}$\rangle$ denotes the molecular SMILES, and $\langle$tag$\rangle$ denotes the classification result. For the BBBP task, we design $\langle$tag$\rangle$ as ``true'' or ``false'', indicating whether the compound can or cannot cross BBB.

Different tasks have different prompts (see Appendix \ref{sec:molnet_prompt}), but we put the tags to the last token of the prompt for all tasks. Let $(s_{i,1},s_{i,2},\cdots,s_{i,T_i})$ denote the $i$-th wrapped sequence for the downstream task with $T_i$ tokens, where $s_{i,T_i}$ is the tag of the sequence. Denote that there are $N$ samples for finetuning. The finetuning strategy could be either 
\begin{equation}
\min -\frac{1}{N}\sum_{i=1}^{N}\log P(s_{i,T_i}|s_{i,<T_i}),
\label{eq:finetune_label_only}
\end{equation}
indicating that we finetune  the tags only, or
\begin{equation}
\min -\frac{1}{N}\sum_{i=1}^{N} \frac{1}{T_i}\sum_{j=1}^{T_i}\log P(s_{i,j}|s_{i,<j}),
\label{eq:finetune_all}
\end{equation}
indicating that we finetune the full prompts. According to our exploration, Eqn.(\ref{eq:finetune_label_only}) achieves slightly better results and we use it for all tasks (see Appendix \ref{sec:moleculenet_detailed_result} for the results).

Let $p_{\rm true}$  and $p_{\rm false}$ denote the probabilities of tags ``true'' and ``false'' after encoding the prefix ``{\em We can conclude that the BBB penetration of }\som{} $\langle$\texttt{SMILES}$\rangle$ \eom{} {\em is}''. 
The probabilities that $\langle$\texttt{SMILES}$\rangle$ can and cannot cross blood-brain barrier are normalized as $p_{\rm true}/(p_{\rm true} + p_{\rm false})$ and $p_{\rm false}/(p_{\rm true} + p_{\rm false})$ respectively. 
The finetuning hyper-parameters are  in Appendix \ref{sec:molnet_hyper}. 

We compare \ourM{} with two types of baselines: (1) pre-trained language model baselines including KV-PLM \cite{Zeng2022KV-PLM}, Galactica~\citep{taylor2022galactica} and MoMu \citep{su2022molecular}. (2) pre-trained Graph Neural Network (GNN) baselines including G-Contextual~\citep{rong2020self}, G-Motif~\citep{rong2020self}, GROVER$_{\rm base}$~\citep{rong2020self}, GROVER$_{\rm large}$~\citep{rong2020self}, GraphMVP \citep{graph_mvp}, MGSSL~\citep{zhang2021motif} and GEM~\citep{fang2022geometry}. The evaluation metric is the ROC-AUC score. The results are in Table \ref{tab:moleculenet}.

\begin{table*}[!htbp]
\centering
\resizebox{0.9\textwidth}{!}{
\begin{tabular}{lcccccccccccccc}
\toprule
{\em Molecule-to-text} & BLEU-2 & BLEU-4 & Rouge-1 & Rouge-2 & Rouge-L  & METEOR  & Text2Mol\\
\midrule
MolT5-small (77M)  & 0.519 & 0.436 & 0.620 & 0.469 & 0.563 & 0.551 & 0.540 \\
MolT5-base (250M) & 0.540 & 0.457 & 0.634 & 0.485 & 0.578 & 0.569 & 0.547 \\
MolT5-Large (800M) & \textbf{0.594} & \textbf{0.508} & 0.654 & 0.510 & 0.594 & 0.614 & 0.582 \\
\midrule
\ourM{} (350M) & \textbf{0.594} & 0.505 & \textbf{0.660} & \textbf{0.511} & \textbf{0.597} & \textbf{0.626} & \textbf{0.594} \\
\midrule
\midrule\textbf{}
{\em Text-to-molecule} &   Exact$\uparrow$  & MACCS$\uparrow$ & RDK$\uparrow$ & Morgan$\uparrow$ & FCD$\downarrow$ & Text2mol$\uparrow$ & Validity$\uparrow$\\
\midrule
MolT5-small  & 0.079  & 0.703 & 0.568 & 0.517 & 2.49 & 0.482 & 0.721 \\
MolT5-medium   & 0.081  & 0.721 & 0.588 & 0.529 & 2.18 & 0.496 & 0.772 \\
MolT5-large   & \textbf{0.311}  & 0.834 & 0.746 & \textbf{0.684} & 1.20&  0.554& 0.905 \\
\midrule
\ourM{} & 0.215  & \textbf{0.859} & \textbf{0.757} & 0.667 & \textbf{0.45} & \textbf{0.578} & \textbf{0.983}\\
\bottomrule
\end{tabular}}
\caption{Results of molecule-to-text (top) and text-to-molecule generation (bottom). For FCD, the smaller, the better. For the remaining metrics, the larger, the better. MolT5 results are from Table 1 and 2 of \cite{molt5}.  MolT5 parameters are from \url{https://github.com/blender-nlp/MolT5}. Bold fonts indicate the best results.
}
\label{tab:text-molecule-translation}
\end{table*}

\ourM{} outperforms the GNN baselines pre-trained on pure molecular data, indicating the effectiveness of pre-training with  scientific text corpus. Compared with Galactica which also uses both SMILES and text for pre-training GPT-like model, \ourM{} obtains better performance. Note that Galactica does not purposely build and train on the ``wrapped'' sequences, whose importance is demonstrated via our empirical results. A possible explanation of the superior performance is that the SMILES describes the component and structural information of molecules, while the text describes the general properties. They are complementary to each other, and joint training on them brings more effective representations.

\subsection{Results on text-molecule translation}
\label{sec:textmol_trans}
We evaluated the performance of \ourM{} on CheBI-20 \cite{text2mol}, a bidirectional text-molecule translation dataset. It consists of 33,010 molecule-description pairs. We use the data split provided by MolT5 \cite{molt5}, where the training, validation and test sets account 80\%, 10\% and 10\% of total data. For molecule-to-text generation, given a molecular SMILES $S$, the prompt is: ``{\em The description of} \som{} $S$ \eom{} {\em is: The molecule is}'', followed by the text description of $S$. For  text-to-molecule generation, given a text description $T$, the prompt is: ``$T$. {\em The compound is} \som'', and the model will generate the molecular SMILES ended with \eom{}. We compare our method with MolT5 \cite{molt5}. 

For molecule-to-text generation, the results are evaluated by NLP metrics including BLEU \cite{bleu}, Rouge~\cite{rouge} and METEOR~\cite{meteor}. ``Text2mol'' is a deep learning based metric proposed by \citet{molt5} to measure the similarity of the text-molecule pairs. For text-to-molecule generation, we evaluate the following metrics: the proportion of the generated SMILES that exactly match the reference SMILES (denoted as ``Exact''); the Tanimoto similarity of three types of fingerprints: MACCS \cite{maccs}, RDK \cite{rdk} and Morgan \cite{morgan}; the FCD score \cite{fcd}, which measures the molecule distances by a pre-trained model; the percentage of the valid generated SMILES. The results are reported in Table \ref{tab:text-molecule-translation}. 

We observe that \ourM{} achieves significantly better performance than MolT5-small and MolT5-base, and has comparable performance with MolT5-large. Note that MolT5-large has $800M$ parameters while \ourM{} only uses 44\% of its parameters. For both tasks, our model performs the best on Text2Mol metric, indicating that \ourM{} captures the alignment between text and molecule better. We attribute it to the wrapped sequences, by which the model can learn the relation between molecule and text explicitly. 

We further verify the zero-shot text-to-molecule generation ability of \ourM{}. The pre-trained \ourM{} takes the text as input and directly generates molecules without finetuning. The top-1 and top-5 fingerprint similarity is in Table \ref{tab:zeroshot_text2drug}. Indeed, compared with the full data setting, the performance drops, but still  reasonable numbers. In addition, the zero-shot \ourM{} successfully recovers $33$ molecules based on the text (see Appendix \ref{sec:examples}). 

\begin{table}[!htbp]
\centering
\resizebox{\linewidth}{!}{
\begin{tabular}{lcccc}
\toprule
&  MACCS & RDK & Morgan\\
\midrule
Zero-shot (Top-1) &   0.540 & 0.383 & 0.228 \\
Zero-shot (Top-5) &  0.580 & 0.423 &  0.423 \\
Full data (Top-1) &  0.841 & 0.746 & 0.660 \\
\bottomrule
\end{tabular}}
\caption{Zero-shot text-to-molecule generation.}
\label{tab:zeroshot_text2drug}
\end{table}

\section{Conclusions and Future Work}
We propose \ourM{},  a generative model pre-trained on scientific text, molecular SMILES and their wrapped sequences. We train a 24-layer \ourM{} with 350M parameters. By prompt-based finetuning, it improves strong baselines on MoleculeNet and achieves comparable results with the best model on molecule-text translation but using much fewer parameters.

For future work, first, we will train larger \ourM{} to further verify the performances across different tasks and the zero-shot/in-context \citep{xie2022an} learning ability. Second, how to further enhance the interaction between molecules and text (e.g., using contrastive learning to enhance consistency) should be studied. Third, how to effectively adapt \ourM{} into other molecule and text tasks such as text-guided molecule optimization is another direction to explore.

\bibliography{custom}
\bibliographystyle{acl_natbib}

\appendix

\noindent\textbf{Appendix}

\section{Datasets and Baselines of MoleculeNet}
\label{sec:moleculenet_baseline}
We choose the following tasks of MoleculeNet for evaluation:

\noindent(1) BBBP contains compounds with binary labels on blood-brain barrier penetration.

\noindent(2) Tox21 is a dataset for predicting the human toxicity of compounds on 12 different targets.

\noindent(3) ClinTox contains drugs approved by the FDA and those that have failed clinical trials for toxicity reasons.

\noindent(4) HIV aims to predict whether a drug can inhibit HIV replication.

\noindent(5) BACE describes binding results for a set of inhibitors of human $\beta$-secretase 1.

\noindent(6) SIDER has compounds used in marketed medicines with 27 categories of side effects. 

\noindent We compare \ourM{} with the following baselines:

\noindent(1) GROVER is a self-supervised pre-trained graph Transformer model. G-Contextual and G-Motif are two variants of it pre-trained with contextual property prediction task and motif prediction task.

\noindent(2) GraphMVP is a self-supervised pre-trained GNN model using both 2D topological structures and 3D geometric views of molecules.
    
\noindent(3) MGSSL leverages a
retrosynthesis-based algorithm BRICS and additional rules to find the motifs and combines motif layers with atom layers.

\noindent(4) GEM is a geometry-enhanced pre-trained GNN model.
    
\noindent(5) {Galactica} is a GPT-like model trained on a large scientific corpus and many natural sequences like SMILES. We report the result of Galactica-120B.

\noindent(6) {KV-PLM} is a BERT-like model where SMILES sequences are appended
after molecule names for pre-training.

\noindent(7) MoMu uses contrastive learning to jointly pre-train a BERT model for text and a GNN model for molecules.

\section{Pre-training hyper-parameters}
\label{app:pretrain_details}
\ourM{} is pre-trained for 200k steps on eight A100 GPUs. The batchsize is 2048 tokens per GPU. The gradients are accumulated for 16 steps before updating. We use Adam~\cite{adam} optimizer for optimization. The  peak learning rate is $0.0005$ and the warm-up steps are 20000. The learning rate scheduler is inverse square root decay scheduler. The dropout is $0.1$.

\begin{figure*}[!htbp] 
\centering
\includegraphics[width=0.85\textwidth]{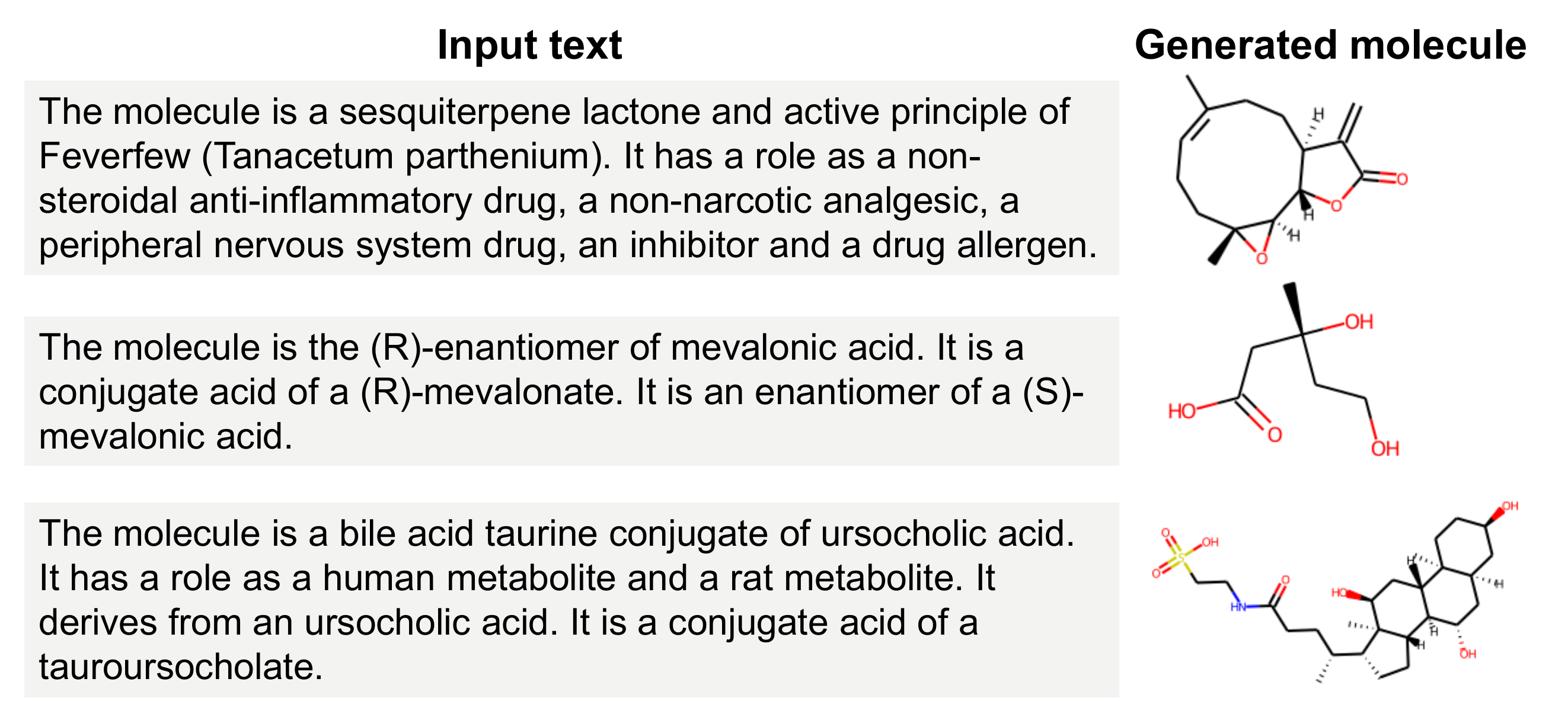}
\caption{Examples for zero-shot text-to-molecule generation. We randomly pick up three cases that \ourM{} can successfully generate the reference molecules without finetuning.}
\label{cases}
\end{figure*}

\section{Finetuning details of downstream tasks}
\label{sec:implementation}
\subsection{Prompts for finetuning MoleculeNet}
\label{sec:molnet_prompt}
\noindent{(1) BBBP}: ``{\em We can conclude that the BBB penetration of }\som{} $\langle$\texttt{SMILES}$\rangle$ \eom{} {\em is true/false.}''

\noindent{(2) Tox21}: ``{\em We can conclude that the }\som{} $\langle$\texttt{SMILES}$\rangle$ \eom{} {\em activity outcome on $\langle$target$\rangle$ is active/inactive.} '' where {\em $\langle$target$\rangle$} refers to corresponding receptor or enzyme for each subtask, e.g. the {\em $\langle$target$\rangle$} of subtask "AR" is "Androgen Receptor".

\noindent{(3) ClinTox}:``{\em We can conclude that the clinical trial toxicity of }\som{} $\langle$\texttt{SMILES}$\rangle$ \eom{} {\em is true/false.}'' for subtask CT\_TOX and ``{\em We can conclude that the FDA approval status of }\som{} $\langle$\texttt{SMILES}$\rangle$ \eom{} {\em is true/false.}'' for subtask FDA\_APPROVED.

\noindent{(4) HIV}: ``{\em We can conclude that the screening result of ability to inhibit HIV replication of }\som{} $\langle$\texttt{SMILES}$\rangle$ \eom{} {\em is active/inactive.}''

\noindent{(5) BACE}: ``{\em We can conclude that the binding result on beta-secretase 1 of }\som{} $\langle$\texttt{SMILES}$\rangle$ \eom{} {\em is true/false.}''

\noindent{(6) SIDER}:``{\em We can conclude that the }\som{} $\langle$\texttt{SMILES}$\rangle$ \eom{} {\em can bring about the side effect of $\langle$side-effect$\rangle$ is true/false.}'' where {\em $\langle$side-effect$\rangle$} refers to corresponding side-effect for each subtask.

\subsection{Details of finetuning MoleculeNet}
\label{sec:molnet_hyper}
We grid search the following hyper-parameters: learning rate in $\{3\times10^{-5},5\times10^{-5}\}$; dropout in $\{0.1,0.3\}$; total epochs from $\{30,50\}$. The model is selected according to validation performance.

\subsection{Details of finetuning text-molecule generation}
For text-molecule generation, \ourM{} is finetuned for 100 steps on one P40 GPU with 1024 tokens and 16 accumulated steps per device. Models are finetuned for 100 epochs.
The learning rate is 0.0001 and the dropout rate is grid searched from $[0.1, 0.2, 0.3, 0.4, 0.5]$. Setting dropout rate as 0.4 and 0.5 achieves the best validation performance on molecule-to-text generation and text-to-molecule generation respectively. We use the corresponding models for testing.

\subsection{MoleculeNet finetuning strategy selection}
\label{sec:moleculenet_detailed_result}
We provide two finetune strategies in  Eqn.\eqref{eq:finetune_label_only} and Eqn.\eqref{eq:finetune_all}. Their results are reported in Table~\ref{tab:moleculenet_detailed}. Their results are similar and Eqn.\eqref{eq:finetune_label_only} is slightly better.

\begin{table*}[!htbp]
\centering
\resizebox{0.9\textwidth}{!}{
\begin{tabular}{lcccccccccccccc}
\toprule
\textbf{Dataset} & BBBP & Tox21 & ClinTox & HIV & BACE & SIDER & Avg \\
\midrule
Dev$_{\rm full~prompt}$ & $98.8\pm0.2$ & $78.8\pm0.1$ & $98.8\pm0.1$ & $82.9\pm1.0$ & $78.4\pm0.3$ & $67.7\pm0.7$ & $84.2$\\
Dev$_{\rm tags~only}$ & $98.9\pm0.3$ & $78.8\pm0.2$ & $97.7\pm0.1$ & $85.3\pm0.2$ & $75.8\pm0.8$ & $69.4\pm0.6$ & $84.3$ \\
\midrule
Test$_{\rm full~prompt}$ & $78.1\pm0.4$ & $77.2\pm0.1$ & $93.4\pm0.1$ & $78.1\pm0.9$ & $87.9\pm0.3$ & $70.0\pm0.2$ & $80.8$\\
Test$_{\rm tags~only}$ & $80.0\pm0.5$ & $77.1\pm0.2$ & $95.3\pm0.2$ & $78.1\pm0.4$ & $88.4\pm1.0$ & $71.7\pm0.2$ & $81.9$ \\
\bottomrule
\end{tabular}}
\caption{Comparison of different finetuning strategies on MoleculeNet. ``Dev'' and ``Test'' denote validation set and test set respectively. Subscripts represent finetuning full prompts (Eqn.\eqref{eq:finetune_all}) or tags only respectively (Eqn.\eqref{eq:finetune_label_only}). The evaluation metric is ROC-AUC. }
\label{tab:moleculenet_detailed}
\end{table*}

\section{Zero-shot text-to-molecule generation} 
\label{sec:examples}
Given $K$ generated molecule $\hat{m}_1,\hat{m}_2,\cdots,\hat{m}_K$ and the reference molecule $m$, the top-$K$ fingerprint similarity is 
\begin{equation}
\max_{i\in[K]} \texttt{similarity}(m,\hat{m}_i).
\end{equation}
\ourM{} generates $33$ molecules that can exactly match the reference molecules without finetuning. Figure \ref{cases} shows three of the cases.

\end{document}